\documentclass[10pt,twocolumn,letterpaper]{article}

\usepackage{iccv}
\usepackage{times}
\usepackage{epsfig}
\usepackage{graphicx}
\usepackage{amsmath}
\usepackage{amssymb}
\usepackage{booktabs}

\usepackage[breaklinks=true,bookmarks=false]{hyperref}
\hypersetup{hidelinks}
\iccvfinalcopy 

\def\iccvPaperID{****} 

\ificcvfinal\fi

\begin{document}

\title{Divide and Ensemble: Progressively Learning for the Unknown}
\author{Hu Zhang~\textsuperscript{1}               \quad
        Xin Shen~\textsuperscript{1}               \quad
        Heming Du~\textsuperscript{1}              \quad
        Huiqiang Chen~\textsuperscript{2}          \quad
        Chen Liu~\textsuperscript{1}               \\
        Hongwei Sheng~\textsuperscript{1}          \quad
        Qingzheng Xu~\textsuperscript{1}           \quad
        MD Wahiduzzaman Khan~\textsuperscript{1}   \quad
        Qingtao Yu~\textsuperscript{1}             \\
        Tianqing Zhu~\textsuperscript{2}            \quad 
        Scott Chapman~\textsuperscript{1}           \quad
        Zi Huang~\textsuperscript{1}                \quad
        Xin Yu~\textsuperscript{1}                  \\
        \textsuperscript{1}University of Queensland, Australia \\
        \textsuperscript{2}University of Technology Sydney, Australia \\
        \texttt{\small \{hu.zhang, xin.yu\}@uq.edu.au}
}


\maketitle
\ificcvfinal\thispagestyle{empty}\fi

\begin{abstract}
   In the wheat nutrient deficiencies classification challenge, we present the DividE and EnseMble (DEEM) method for progressive test data predictions.
   We find that (1) test images are provided in the challenge; (2) samples are equipped with their collection dates; (3) the samples of different dates show notable discrepancies. Based on the findings, we partition the dataset into discrete groups by the dates and train models on each divided group. We then adopt the pseudo-labeling approach to label the test data and incorporate those with high confidence into the training set. In pseudo-labeling, we leverage models ensemble with different architectures to enhance the reliability of predictions. The pseudo-labeling and ensembled model training are iteratively conducted until all test samples are labeled. Finally, the separated models for each group are unified to obtain the model for the whole dataset. Our method achieves an average of 93.6\% Top-1 test accuracy~(94.0\% on WW2020 and 93.2\% on WR2021) and wins the 1$st$ place in the Deep Nutrient Deficiency Challenge~\footnote{https://cvppa2023.github.io/challenges/}.
\end{abstract}

\section{Introduction}
Precise fertilization to crops is essential for sustainable agriculture, \emph{e.g.}, maximizing crop yields while reducing the negative environmental impacts of fertilizers~\cite{yi4549653non,heinemann2021simplifying,berti2016overview}. To realize this, accurate soil nutrient assessments are imperative. In light of this, a novel dataset tailored for wheat nutrition deficiency classification has been introduced, diverging significantly from conventional ones predominantly focused on object recognition~\cite{deng2009imagenet}. This initiates a unique classification task that demands accurate predictions of soil nutrient conditions through nuanced variations in crops, which is different from existing tasks~\cite{krizhevsky2012imagenet, zhang2023divide}. The intrinsic challenges within this novel dataset necessitate the development of innovative algorithms or models.

\begin{figure}
\begin{center}
\includegraphics[width=0.9\linewidth]{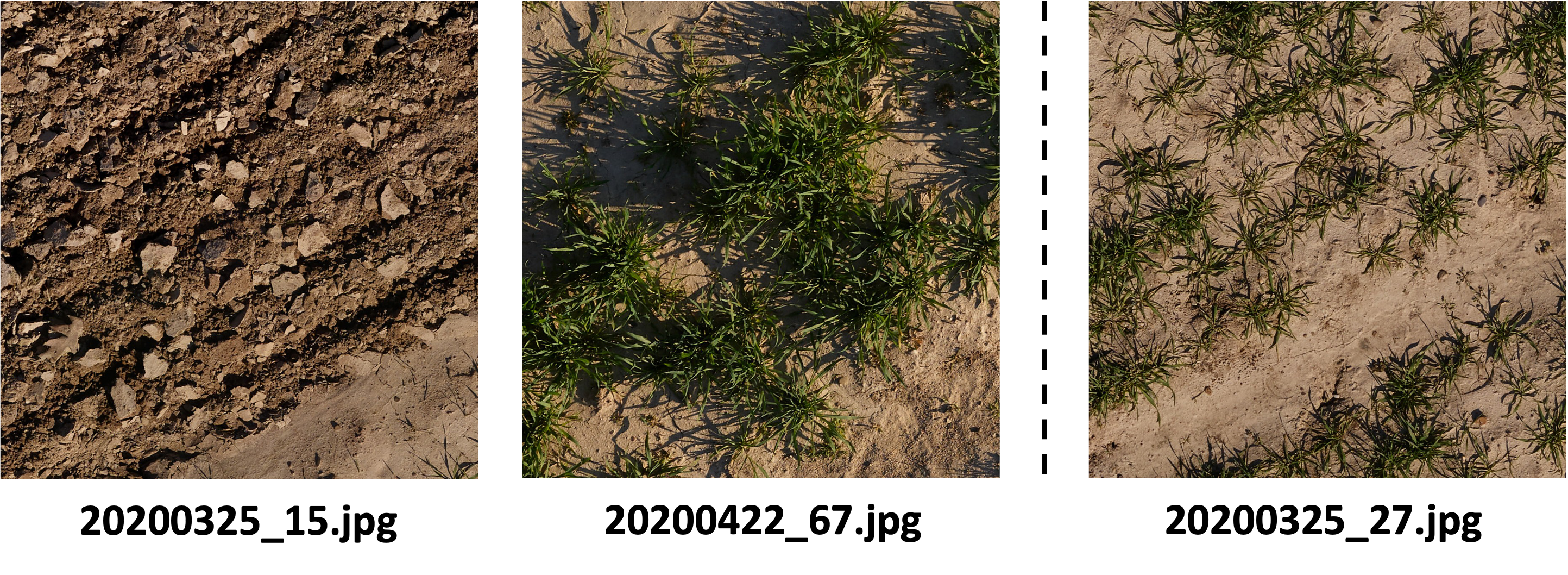}
\end{center}
   \caption{The first two images originate from the training set, collected on varying dates but have the same label: unfertilized. The last image is from the test set with the collection date in the name.}
\label{fig:intro}
\end{figure}

In this report, we propose the Divide and Ensemble strategy (DEEM) to progressively predict the labels of testing data. We observe that each sample is named with its corresponding collection date and the samples with different dates exhibit significant differences, indicating the domain gaps between them (see Figure~\ref{fig:intro}). To mitigate such gaps and facilitate effective model training, we first split a dataset based on the dates and deal with each split separately.

Initially, we employ the original training samples to obtain the preliminary model. Leveraging this model and the testing images available in the challenge, we progressively allocate pseudo-labels to test samples, subsequently integrating these labeled samples into the iterative training paradigm~\cite{lee2013pseudo}. This strategy ensures a phased inclusion of the test samples into the model, progressing from easy to hard ones. To mitigate the potential risk of overfitting or bias inherent in a single model, we adopt the idea of the models ensemble in pseudo-labeling, a well-recognized strategy for improving performance across a broad spectrum of tasks~\cite{ganaie2022ensemble}. Specifically, we utilize a variety of models characterized by distinct architectures, such as ResNet and EfficientNet, each operating as a specialized expert in tackling the given classification problem. These ``experts'' collaborate in determining the most plausible pseudo-label for each test sample through majority voting criteria: All experts have the same prediction or only one expert is diverged in prediction. The rest test samples will not be assigned pseudo-labels in this round. This strict criterion ensures a high probability of label correctness.
\begin{figure*}
\begin{center}
\includegraphics[width=0.98\linewidth]{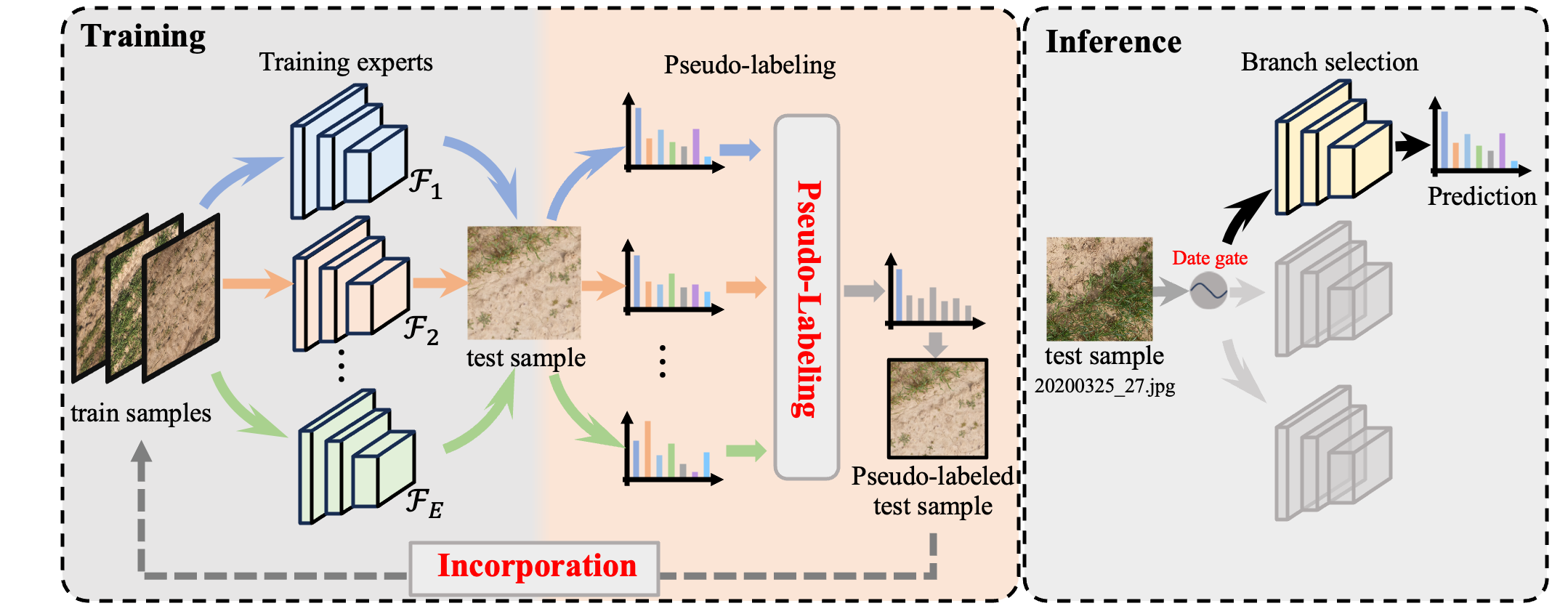}
\end{center}
   \caption{The overall framework of our proposed method. For each grouped data after date-based splitting, we train multiple experts and gradually assign pseudo-labels to testing samples. The labeled testing samples are incorporated into the original training set and update multi-experts iteratively. The model in each grouped data are treated as one branch for the whole dataset and are unified together. In the inference, the collection date of the testing sample is used to select the appropriate branch for feature extraction and class prediction.}
\label{fig:overall framework}
\end{figure*}
Incorporating these accurately labeled test samples into the training set allows for the constant refinement of the ensembled models, resulting in an iterative label assignment for the entire test dataset and ensuring optimal predictions for each split. Finally, we unify the models from different splits to obtain the final model for the whole dataset.

In the experiments, we obtained an impressive average accuracy of 93.6\% in the given test data, with scores of 94.0\% and 93.2\% for WW2020 and WR2021, respectively. Further details of the DEEM framework and an analysis of some key factors are explored in Section~\ref{sec: method} and~\ref{sec: experiments}.

\section{Method}
\label{sec: method}
In this section, we present the details of our method. It consists of three major components: Dataset splitting, pseudo-labeling with the models ensemble, and the incorporation of test samples followed by new iterative training. The overall framework is shown in Figure~\ref{fig:overall framework}.

\subsection{Date-based Splitting and Modeling}
Suppose the dataset is denoted as $\mathcal{D} = [\mathcal{D}_{train}, \mathcal{D}_{test}]$. $\mathcal{D}_{train} = \{(x_i, a_i, y_i)\}_{i=0}^{N_{train}}$, where $x_i$ is the RGB image, $a_i$ is the accompanied names with the structure of ``date\_sequence.jpg'' and $y_i$ is the corresponding label. $N_{train}$ denotes the number of training samples in the dataset. $\mathcal{D}_{test} = \{(x_i, a_i)\}_{i=0}^{N_{test}}$ is the testing set with only the RGB images and the image names. 

Based on the image name $a_i$, we extract the ``date'' attribute from each sample and derive a date set defined as $\mathcal{S} = (date_1, ..., date_M)$. Leveraging this set $\mathcal{S}$, we group the training and testing samples with the same collection date, thereby constituting the newly grouped dataset $\mathcal{D}_m = [\mathcal{D}_{train}^m, \mathcal{D}_{test}^m], m=1,..., M$.

Rather than train the model directly on the whole dataset, we instead train models $f$ with the training samples in each grouped dataset $\mathcal{D}_m$. Predictions for the testing samples are thus separately made in each group. For simplicity, we consider one grouped dataset for illustration and adopt the same notation as the whole dataset $\mathcal{D} = [\{(x_i, a_i, y_i)\}_{i=0}^{N_{train}}, \{(x_i, a_i)\}_{i=0}^{N_{test}}]$.

\subsection{Pseudo-Labeling with Models Ensemble}
\label{sec: Pseudo-Labeling with Models Ensemble}
To predict the labels of testing samples in $\mathcal{D}_{test} = \{(x_i, a_i)\}_{i=0}^{N_{test}}$, we consider models with different architectures such as ResNet50 and ResNext50, which are trained utilizing the training set $\mathcal{D}_{train} = \{(x_i, a_i, y_i)\}_{i=0}^{N_{train}}$. The obtained initial models are denoted as $\mathcal{F}_0, \cdots, \mathcal{F}_E$ and each of the models is a specialized expert in addressing the nutrient deficiencies problem. Rather than predicting the final labels to all the testing samples directly, we progressively assign pseudo-labels to the test samples, incorporating them into an iterative training paradigm.

Take one test sample $x_i$ as an example. Each expert $\mathcal{F}_e$ outputs the probability distribution $p_i^e = \mathcal{F}_e(x_i)$. The potential label $\hat{y}_i^{e}$ is the class with the largest value in $p_i^e$. We thus obtain a set of predicted labels from all models, denoted as $\{\hat{y}^e_i, e=1,..., E\}$. The pseudo-label of $x_i$ is finally determined based on the criteria outlined below:

\noindent \textbf{Case 1.} Consistent predictions are observed across different experts $\mathcal{F}_0, \cdots, \mathcal{F}_E$, resulting in a uniform labels: $\hat{y}^0_i = \hat{y}^1_i\cdots = \hat{y}^E_i$. This scenario suggests a stable and robust prediction for sample $x_i$. Assuming the number of experts is sufficiently large, this label is likely to be accurate. Thus, we directly assign this label to $x_i$, treating it as the true label. This process results in the labeled testing set $\mathcal{D}_{test}^{1} = \{(x_i, a_i, \hat{y}_i)\}_{i=0}^{N_{test}^1}$ when considering different samples together, where $N_{test}^1$ denotes the number of labeled samples in this case and $\hat{y}_i$ is the assigned pseudo-label.

\noindent \textbf{Case 2.} Discrepancies emerge in the predictions made by experts $\mathcal{F}_0, \cdots, \mathcal{F}_E$. The samples with different predictions are intrinsically challenging to classify, lying near the boundaries of distinct categories. The ``hard samples'' may be correctly classified by one model while being misclassified by another, highlighting their critical role in enhancing model performance.

In this case, it is noted that despite the divergence in Top-1 predictions, Top-2 predictions are often aligned. For instance, one expert predicts $(5,4)$ while another expert predicts $(4,5)$, suggesting a narrow range where the true label likely falls. If $E-1$ in $E$ experts predict exactly the same Top-2 labels, we then select the most frequently occurring label as the potential pseudo-label. To further substantiate the correctness of predicted labels, a feature similarity analysis is employed. It requires the extraction of features from the test sample $x_i$ and all samples in the training set $\mathcal{D}_{train}$. Then, the cosine similarities between testing and training features are calculated. The Top-k training instances with the least distances are selected and their labels are collected. This process enables the validation of whether the most occurring label in the Top-k aligns with the potential pseudo-label above, confirming the label consistency in both probability and feature domains.

This strict method promises a highly accurate label assignment. Considering all test samples together, we obtain the labeled samples $\mathcal{D}_{test}^{2} = \{(x_i, a_i, \hat{y}_i)\}_{i=0}^{N_{test}^2}$, where $N_{test}^2$ denotes the number of labeled samples in this case.

\noindent \textbf{Case 3.} In scenarios where the above conditions are unfulfilled, no label is assigned in the current iteration.

\subsection{Test Samples Incorporation}
Following this, we use the newly labeled samples from $\mathcal{D}_{test}^{1}$ and $\mathcal{D}_{test}^{2}$ to update the training dataset $\mathcal{D}_{train}$, creating $\mathcal{D}_{train}^{update}$. We then retrain the ensembled models $\mathcal{F}_0, \cdots, \mathcal{F}_E$ on this updated dataset by following the previous training protocols. This process is repeated iteratively to gradually assign pseudo-labels to all test data.

\subsection{Final Model and Inference}
After assigning pseudo-labels to all test data, we train a final model for each data split, saving the last epoch checkpoint. The model for the whole dataset is structured with branch models from different splits, chosen based on the sample date. In inference, the test sample name is input to extract the date, which automatically guides the appropriate branch selection for feature extraction and class prediction.
\section{Experiments}
\label{sec: experiments}
\subsection{Dataset and Evaluation Metric}
The wheat nutrient deficiencies dataset consists of two parts collected in different years, \emph{i.e.}, WW2020, and WR2021. Each part comprises 1332 training images and 468 test images. There are seven categories representing various nutrient deficiency cases: NPKCa+m+s, NPKCa, \_PKCa, N\_KCa,
NP\_Ca, NPK\_, and unfertilized. The objective is to accurately predict the test sample labels utilizing the data at hand. The performance metric employed in this challenge is the Top-1 accuracy, measured across both test sets derived from the two datasets.
\subsection{Implementation Details}
Our method is grounded on the given codebase. We split both WW2020 and WR2021 into three groups based on the distinct dates in each part. Specifically, we obtain group ``20200314'', ``20200422'', ``20200506'' for WW2020, ``20200314'', ``20210314'', ``20210506'' for WR2021. Each group evenly consists of a third of the total training and testing samples. For the model training on each group, we adopt ResNet-50, ResNet-101, ResNext50, and EfficientNet\_v2\_s, thereby setting the expert number ($E$) to four. The images are resized to 1135 and the batch size is 8. The original optimizer SGD is replaced with Adam and the learning rate is set as 1e-3. The weight decay is also set as 1e-3. The scheduler is ``ReduceLROnPlateau'' with patience of 3, and the factor is changed to 0.5. For data augmentation, we adopt the ``RandomRotation(180)'', ``RandomHorizontalFlip()'', ``RandomVerticalFlip()'', mean and variance normalization, and ``RandomErasing()''. The Top-k in Section~\ref{sec: Pseudo-Labeling with Models Ensemble} (Case 2) is set as 10. The above configuration is applied to all models across different splits.

\subsection{Results}
We report our final results and a variety of baselines from work~\cite{yi4549653non} in Table~\ref{tab: main results}. We also include the baseline provided in github~\footnote{\url{https://github.com/jh-yi/DND-Diko-WWWR}\label{github}}. Compared with the best-performing baseline which achieves 84.9\% in Top-1 accuracy, our DEEM method attains 93.6\% Top-1 accuracy on the test, which is 8.7\% higher. More specifically, we achieved 10.5\% higher in WW2020 and 6.9\% higher in WR2021.

\begin{table}
\begin{center}
\begin{tabular}{l|ccc}
\toprule
Method & WW2020 & WR2021 & Average\\
\midrule
Swin v1~\cite{yi4549653non}  & 80.6 & 83.8 & 82.2 \\
DenseNet-161~\cite{yi4549653non}  & 78.8 & 87.0 & 82.9 \\
Swin v2~\cite{yi4549653non}  & 83.5 & 86.3 &84.9 \\
Swin\_v2\_s~\textsuperscript{\ref{github}}  & 79.5 & 84.0 &81.7 \\
\midrule 
DEEM (Ours) & 94.0 & 93.2 & 93.6 \\
\bottomrule
\end{tabular}
\end{center}
\caption{Top-1 accuracy (\%) on the test set. }
\label{tab: main results}
\end{table}
\subsection{Ablation Study}
In this part, we choose WW2020 and study different components of our method. We randomly reserved 200 samples from the 1332 training samples for validation.

\noindent \textbf{Separation or Unification.}
We first study the benefits of splitting the dataset based on the collected date of each sample. We adopt the RestNet50 as the backbone and train individual models for three groups. We also train one model on the whole WW2020 dataset. The performance on validation samples is detailed in Table~\ref{tab: Separation or Unification}.

Take the ``20200314'' group as an example, the method trained on separate groups exhibits a higher validation accuracy, increasing from 85.7\% to 90.9\%. This indicates that more data does not necessarily bring gains in performance and the domain gaps between the samples collected on different dates will hurt the training of models.

\begin{table}
\begin{center}
\scriptsize
\begin{tabular}{l|cccc}
\toprule
Split & 20200314 & 202000422 & 20200506 & Average\\
\midrule 
Whole model & 85.7 &90.5 &93.8 & 90.0 \\
\midrule
Individual model & 90.9&  93.7&  96.7 & 93.8\\
\bottomrule
\end{tabular}
\end{center}
\caption{Top-1 accuracy (\%) between individual models and whole model on the validation set. }
\label{tab: Separation or Unification}
\end{table}
\noindent \textbf{Direct voting and progressive learning.}
We investigate the impact of the progressive learning approach with pseudo-labels. Specifically, we first directly predict the labels for all the validation data with model ensembles (Direct voting). We then adopt the pseudo-labeling strategy in this report and assign labels to the validation data. We gradually incorporate the labeled samples into training until all validation samples are labeled (Progressive learning).

The results of two cases are recorded in Table~\ref{tab: progressive learning}. The progressive learning strategy consistently achieves better results across all splits. For example, the ``PL'' obtains 99.6\% in accuracy, which is 5.8\% higher than direct voting. Such results validate the effectiveness of our designed progressive pseudo-labeling strategy.

\begin{table}
\scriptsize
\begin{center}
\begin{tabular}{l|cccc}
\toprule
Split & 20200314 & 202000422 & 20200506 & Average\\
\midrule
DV &90.9  &93.7  &96.7 &93.8 \\
\midrule 
PL &98.7  & 100.0 &100.0 & 99.6\\
\bottomrule
\end{tabular}
\end{center}
\caption{Top-1 accuracy (\%) between ``direct voting (DV)'' and ``progressive learning (PL)'' on the validation set.}
\label{tab: progressive learning}
\end{table}
\noindent \textbf{Number of Experts.}
We study the varying number of experts on the final performance. Specifically, we adopt 1-5 experts from \{ResNet50, ResNext50, ResNet101, EfficientNet\_v2\_s, Regnet\_y\_400mf\} in sequence to perform the progressive pseudo-labeling. The results are shown in Table~\ref{tab: Number of Experts}.

We observe that with the increased number of experts, the results of the validation samples are improved from 94.2\% to 99.6\%. The performance remains stable when the number is large enough. Thus, four experts were chosen in our experiments.

\begin{table}
\scriptsize
\begin{center}
\begin{tabular}{l|cccc}
\toprule
Number & 20200314 & 202000422 & 20200506 & Average\\
\midrule
1 &92.2  &93.7  &96.7 &94.2\\
\midrule 
2 &96.1  &96.8 &100.0 &97.6\\
\midrule 
3 &98.7  & 98.4 &100.0 & 99.0\\
\midrule 
4 &98.7  & 100.0 &100.0 & 99.6\\
\midrule 
5 &98.7  & 100.0 &100.0 & 99.6\\
\bottomrule
\end{tabular}
\end{center}
\caption{Top-1 accuracy (\%) with different number of experts.}
\label{tab: Number of Experts}
\end{table}
\section{Conclusion}
In this study, we introduce the DEEM approach for the wheat nutrient deficiencies challenge. Initially, we split the dataset into distinct groups, facilitating dedicated model training in each group. Subsequently, we employ a model ensemble strategy for test samples pseudo-labeling, followed by the iterative inclusion of labeled test samples and a new round of model training. In the experiment, our method achieves an average of 93.6\% Top-1 accuracy on the test set. We hope that our work can provide some insights into future work in this area.

\noindent 
\noindent \textbf{Acknowledgements.}
This research is funded in part by ARC-Discovery grant (DP220100800 to XY) and ARC-DECRA grant (DE230100477 to XY).
{\small
\bibliographystyle{ieee_fullname}
\bibliography{egbib}

\begin{thebibliography}{1}\itemsep=-1pt

\bibitem{berti2016overview}
Antonio Berti, Anna Dalla~Marta, Marco Mazzoncini, and Francesco Tei.
\newblock An overview on long-term agro-ecosystem experiments: Present
  situation and future potential.
\newblock {\em European Journal of Agronomy}, 77:236--241, 2016.

\bibitem{deng2009imagenet}
Jia Deng, Wei Dong, Richard Socher, Li-Jia Li, Kai Li, and Li Fei-Fei.
\newblock Imagenet: A large-scale hierarchical image database.
\newblock In {\em 2009 IEEE conference on computer vision and pattern
  recognition}, pages 248--255. Ieee, 2009.

\bibitem{ganaie2022ensemble}
Mudasir~A Ganaie, Minghui Hu, AK Malik, M Tanveer, and PN Suganthan.
\newblock Ensemble deep learning: A review.
\newblock {\em Engineering Applications of Artificial Intelligence},
  115:105151, 2022.

\bibitem{heinemann2021simplifying}
Paul Heinemann and Urs Schmidhalter.
\newblock Simplifying residual nitrogen (nmin) sampling strategies and crop
  response.
\newblock {\em European Journal of Agronomy}, 130:126369, 2021.

\bibitem{krizhevsky2012imagenet}
Alex Krizhevsky, Ilya Sutskever, and Geoffrey~E Hinton.
\newblock Imagenet classification with deep convolutional neural networks.
\newblock {\em Advances in neural information processing systems}, 25, 2012.

\bibitem{lee2013pseudo}
Dong-Hyun Lee et~al.
\newblock Pseudo-label: The simple and efficient semi-supervised learning
  method for deep neural networks.
\newblock In {\em Workshop on challenges in representation learning, ICML},
  volume~3, page 896. Atlanta, 2013.

\bibitem{yi4549653non}
Jinhui Yi, Gina Lopez, Sofia Hadir, Jan Weyler, Lasse Klingbeil, Marion
  Deichmann, Juergen Gall, and Sabine~J Seidel.
\newblock Non-invasive diagnosis of nutrient deficiencies in winter wheat and
  winter rye using uav-based rgb images.
\newblock {\em Available at SSRN 4549653}.

\bibitem{zhang2023divide}
Hu Zhang, Linchao Zhu, Xiaohan Wang, and Yi Yang.
\newblock Divide and retain: A dual-phase modeling for long-tailed visual
  recognition.
\newblock {\em IEEE Transactions on Neural Networks and Learning Systems},
  2023.

\end{thebibliography}
}

\end{document}